\documentclass[11pt]{article}

\usepackage[final]{acl}

\usepackage{times}
\usepackage{latexsym}
\usepackage[T1]{fontenc}
\usepackage[utf8]{inputenc}
\usepackage{microtype}
\usepackage{inconsolata}
\usepackage{graphicx}
\usepackage{makecell}
\usepackage{subcaption}
\usepackage{xspace}
\usepackage{CJKutf8}
\usepackage{booktabs}
\usepackage{multirow}
\usepackage{longtable}
\usepackage{tabularx}
\usepackage{pifont}
\usepackage{float}             
\usepackage{tcolorbox}        
\tcbset{
  colback=white,              
  colframe=black,             
  boxrule=0.8pt,              
  arc=2mm,                    
  left=4pt,right=4pt,top=4pt,bottom=4pt  
}

\def\SampleNum{4,870\xspace}
\def\LangNum{10\xspace}
\def\CountryNum{10\xspace}
\def\RegionNum{18\xspace}
\def\TaskNum{10\xspace}
\def\BenchName{CulturalMenuBench\xspace}

\title{\BenchName: Probing the Knowledge-Application Gap \\in Multimodal Culinary Reasoning}

\author{
  \textbf{Bo Zeng\textsuperscript{1}},
  \textbf{Linfeng Gao\textsuperscript{1,2}},
  \textbf{Peiqin Lin\textsuperscript{1}},
  \textbf{Yu Zhao\textsuperscript{1}} \\
  \textbf{Mingyan Zeng},
  \textbf{Yu Tong},
  \textbf{Xintong Wang\textsuperscript{1,3}\thanks{Correspondence to \nolinkurl{xintong.wang@uni-hamburg.de}.}},
  \textbf{Linlong Xu\textsuperscript{1}} \\
  \textbf{Longyue Wang\textsuperscript{1}},
  \textbf{Weihua Luo\textsuperscript{1}},
  \textbf{Qinggang Zhang\textsuperscript{2}},
  \textbf{Jinsong Su\textsuperscript{2}} \\
  \textsuperscript{1}Alibaba Group \quad
  \textsuperscript{2}Xiamen University \quad
  \textsuperscript{3}University of Hamburg
}

\begin{document}
\maketitle
\begin{abstract}
  Multimodal language models achieve near-ceiling scores on food recognition benchmarks, yet it remains unclear whether this success reflects genuine cultural understanding or mere visual matching. To probe this distinction, we introduce \BenchName, a benchmark of \SampleNum items in \LangNum languages across \RegionNum regions; its \TaskNum tasks pair final-dish and step-by-step cooking images with ingredients, procedural text, and regional labels, spanning basic recognition to process-grounded cultural attribution. Evaluating 12 models exposes a substantial knowledge-application gap: models exceeding 94\% on standard multiple-choice tasks drop to at most 56\% when attributing dishes to Chinese regional cuisines, despite an identical four-way format. Diagnostic analyses explain why: error patterns are consistent with random guessing, accuracy tracks visual distinctiveness rather than cultural structure, and models classify cuisines more accurately from dish names alone than from images (+7--18 points). The knowledge is thus present but cannot be activated through visual input. An ablation confirms these tasks genuinely require procedural evidence: removing sequential cooking images selectively degrades process-grounded tasks while others remain stable. Overall, \BenchName shows that near-perfect recognition can conceal an inability to apply cultural knowledge, motivating training that explicitly connects perception, procedure, and cultural context. Code and data are publicly available.\footnote{\url{https://github.com/BobTsang-NLP/CulturalMenuBench}}
  \end{abstract}

\section{Introduction}

Do Large Language Models (LLMs) truly \emph{understand} culinary culture, or do they merely \emph{discriminate} surface patterns when prompted with explicit candidates? As LLMs are deployed in dietary recommendation, culinary tourism, and cross-cultural education, this distinction between knowledge \emph{possession} and \emph{application} becomes critical~\citep{cao2024exploring, fung2024massively, li2024culturellm}. Genuine culinary understanding demands not only recognizing a dish from its image, but also inferring ingredient--technique associations from multi-step cooking processes and situating these observations within specific cultural and regional contexts~\citep{liu-etal-2021-visually, cao2023cultural, bugliarello-etal-2023-measuring, qi2026industrybench, bai2026industrybench}.

Cuisine offers a fertile testbed for probing cultural competence, yet existing benchmarks~\citep{DBLP:conf/nips/RomeroLWGMPOVBJ24,DBLP:conf/cvpr/VayaniDWASTAHKK25,ma2023food} remain circumscribed in both reasoning depth and cultural coverage (Tab.~\ref{tab:vqa_datasets_modified}). FoodieQA \citep{li2024foodieqa} and WC-VQA \citep{DBLP:conf/naacl/WinataHIAPWNPOARDPAWMLA25} target direct visual recognition and question answering, omitting the cooking processes in which much of culinary understanding is encoded. IndiFoodVQA \citep{agarwal2024indifoodvqa} advances toward compositional evaluation but is confined to English and does not capture intra-country regional distinctions. More critically, these shallow tasks engender a capability illusion: models readily surpass 90\% on candidate matching, masking their true capacity for independent reasoning.

\begin{table}[htbp]
\centering
\small
\resizebox {\columnwidth} {!} {
\begin{tabular}{lccccc}
\toprule
\textbf{Benchmark} & \textbf{Reasoning} & \textbf{\#T} & \textbf{\#L} & \textbf{Food Info} & \textbf{Sub-region} \\
\midrule
FoodieQA                  &  One-hop     & 3     & 1   & \makecell{dish name\\location\\flavor\\dish image} & Yes \\ \hline
IndiFoodVQA               &  Multi-hop   & 1     & 1   & \makecell{dish name\\location\\raw material\\flavor\\dish image} & No \\ \hline
WC-VQA                    &  One-hop     & 2     & 30  & \makecell{dish name\\location\\dish image}         & No \\ \hline
Ours                      &  \makecell{Process-\\grounded}   & 10    & \LangNum  & \makecell{dish name\\location\\raw material\\practice\\dish image\\practice images} & Yes \\
\bottomrule
\end{tabular}
}
\caption{\textbf{Statistics of representative culinary benchmarks.} \#T: number of tasks, \#L: number of languages, Food Info: food information, Sub-region: whether includes sub-region information.}
\label{tab:vqa_datasets_modified}
\end{table}

To address this, we present \BenchName, assembled through a scalable pipeline from web-mined menus. The benchmark offers three key advantages: 1) \textbf{rich process-level metadata}: beyond dish names and final images, it encompasses ingredients, step-by-step instructions, process images, and sub-regional tags; 2) \textbf{reasoning-intensive tasks}: \TaskNum tasks spanning visual recognition, cross-modal matching, process-grounded reasoning, and cultural classification; and 3) \textbf{multi-granularity cultural coverage}: \RegionNum regions across \CountryNum countries in \LangNum languages, with provincial-level labels for Chinese cuisines enabling both cross-cultural and intra-cultural evaluation.

Systematic evaluation of 12 models uncovers a \textbf{knowledge-application
gap}: leading models attain 94--97\% on standard four-way MC tasks, yet at
most 56\% when classifying sub-regional Chinese cuisines under the same
four-way response interface. With the format held constant, this 40--50 point
decline reflects a genuine failure of knowledge application rather than a
response-format effect. Diagnostic error analysis exposes the mechanism: confusion patterns are consistent with random guessing, and per-cuisine accuracy strongly tracks visually distinctive cuisines (Sichuan 78\% vs.\ Anhui 19\%), indicating reliance on visual shortcuts rather than cultural reasoning.

Ablation corroborates that this finding is not an artifact of benchmark design: retaining only the first image per step group, thereby removing the remaining sequential context, selectively degrades process tasks by 5.7--11.7pp while non-process tasks remain stable ($\leq$1.0pp). This double dissociation establishes that the benchmark elicits authentic procedural reasoning, rendering the cultural reasoning failure all the more striking. 

These findings expose a structural deficiency: the reasoning bridge from perception to cultural knowledge is not reliably activated, and this failure mode is consistent across architectures and scales. Our contributions are threefold:

\begin{itemize}
    \item We introduce \BenchName, a multimodal benchmark pairing process-level culinary metadata (step-by-step images, ingredients, procedural text) with sub-regional labels across \LangNum languages and \RegionNum regions.

    \item We devise a scalable pipeline yielding \SampleNum items from web-mined menus, instantiating \TaskNum tasks from visual recognition to process-grounded reasoning.

    \item We uncover a knowledge-application gap, including an approximately 40--50 point decline for leading matched systems under the same four-way response interface, and diagnose its mechanism: models rely on visual shortcuts rather than cultural reasoning, a failure consistent across 11 model/configuration conditions spanning 10 unique checkpoints.
\end{itemize}

\section{\BenchName}

\begin{figure*}[t]
  \centering
  \resizebox{\textwidth}{!}{%
    \includegraphics[width=\linewidth]{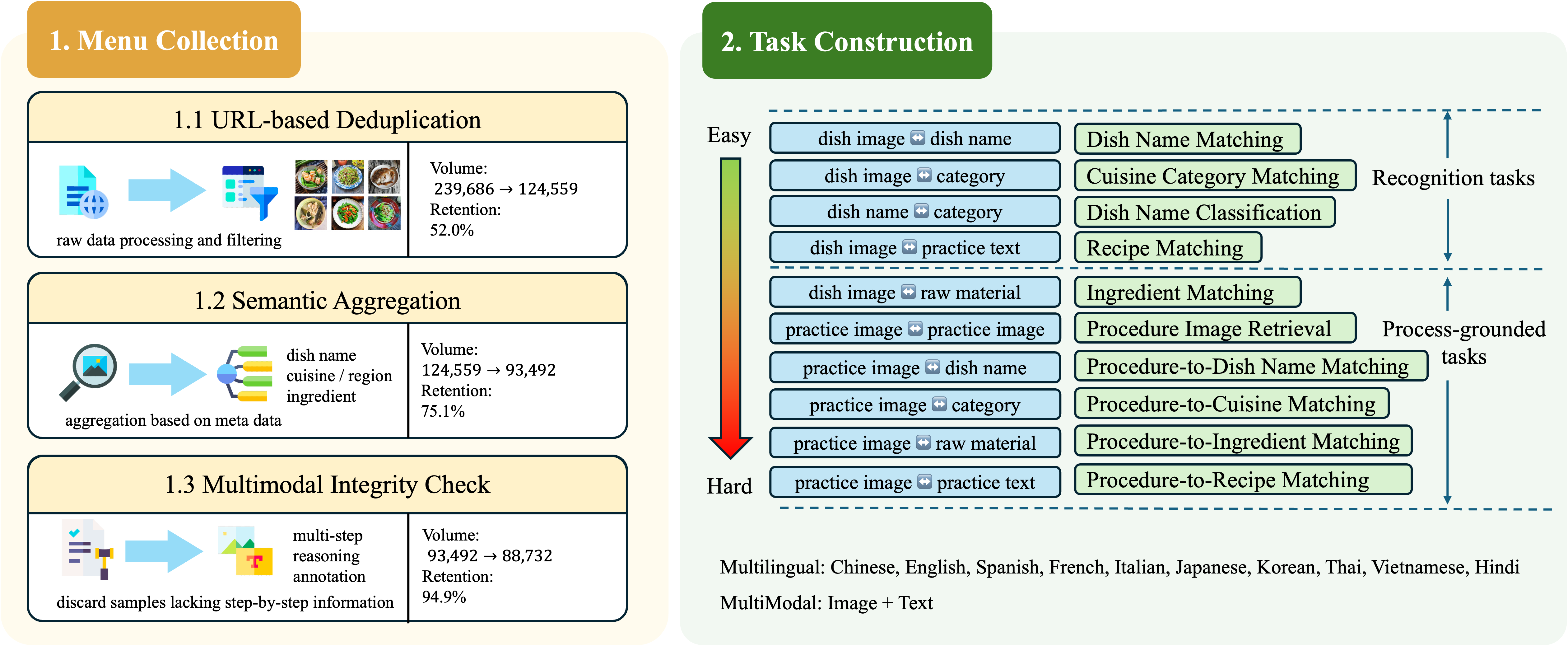}%
  }
  \caption{\textbf{Overview of \BenchName construction.} The pipeline comprises three-stage filtering (URL deduplication, semantic aggregation, integrity checks) from 240K raw records to 587 curated samples, followed by task construction spanning visual recognition to process-grounded cultural reasoning.}
  \label{fig:pipeline}
\end{figure*}

To evaluate the process-grounded reasoning introduced above, \BenchName furnishes each of its 587 curated dishes with step-by-step cooking images, ingredient lists, and procedural text alongside the final dish image, which are necessary to trace technique sequences and intermediate states that distinguish regional traditions. The benchmark comprises \SampleNum evaluation items spanning \LangNum languages and \RegionNum regions. Construction proceeds in two stages (Fig.~\ref{fig:pipeline}): menu collection (\S\ref{sec:data_construction}) and task construction (\S\ref{sec:task_construction}).

\subsection{Menu Collection Pipeline}
\label{sec:data_construction}

\begin{table}[htbp]
\centering
\small
\resizebox{\columnwidth}{!}{
\begin{tabular}{l|r|r}
\toprule
\textbf{Pipeline Stage} & \textbf{Volume} & \textbf{Retention} \\
\midrule
1. Raw Data Crawling & 239,686 & 100.0\% \\
2. URL-based Dedup.  & 124,559 & 52.0\% \\
3. Semantic-level Dedup.  & 93,492 & 75.1\% \\
4. Multimodal Integrity Check & 88,723 & 94.9\% \\
5. \textbf{Final Curation} & \textbf{587} & \textbf{0.7\%} \\
\bottomrule
\end{tabular}
}
\caption{\textbf{Data retention statistics across each stage.} The entire pipeline guarantees the completeness and consistency of core metadata (Dish Image + Practice Image + Textual Description).}
\label{tab:data_funnel}
\end{table}

To curate menus of requisite quality and diversity, we devise a dedicated collection pipeline. We source structured recipe data from MeishiChina\footnote{\url{https://www.meishichina.com/}}, a large-scale Chinese culinary platform hosting user-contributed recipes spanning both domestic Chinese and internationally adopted cuisines (e.g., Japanese, Italian, Thai, Korean). This single-platform sourcing strategy is a deliberate design choice conferring two key advantages: (1)~\textbf{structural homogeneity}: all recipes share a consistent schema of dish images, step-by-step instructions with photos, and ingredient lists, enabling uniform task construction across cuisines; and (2)~\textbf{cross-cultural adaptation as an evaluation target}: the platform's international recipes reflect how dishes are adapted and reinterpreted across cultural boundaries, a phenomenon central to real-world culinary knowledge that models must handle. This unified-source design enables controlled comparison in which the primary variable is model capability, rather than confounding differences in data schema, annotation standards, or image quality across heterogeneous platforms. From the initial 239,686 records, we apply a multi-stage filtering regimen summarized in Tab.~\ref{tab:data_funnel}:

\paragraph{URL-based Deduplication} 
We first eliminate redundant entries by deduplicating on the unique URL associated with each recipe. This stage discards mirrored or near-identical submissions, reducing the corpus to 124,559 distinct records.

\paragraph{Semantic Aggregation}
To address semantic redundancy, where different users submit recipes for the same dish under variant names or descriptions, we encode dish names using a multilingual sentence embedding model\footnote{https://huggingface.co/machimachiop/paraphrase-multilingual-MiniLM-L2-v2} and perform greedy deduplication: items are ranked by a composite quality score (number of practice images + text length), and each candidate is compared against all previously retained items via cosine similarity. A candidate is discarded if its similarity to any retained item exceeds a threshold of 0.9, unless a core-ingredient conflict is detected (e.g., ``braised pork'' vs.\ ``braised fish''). This procedure yields 93,492 semantically unique dish entities.

\begin{figure*}[t]
  \centering
  \resizebox{0.9\textwidth}{!}{%
  \includegraphics{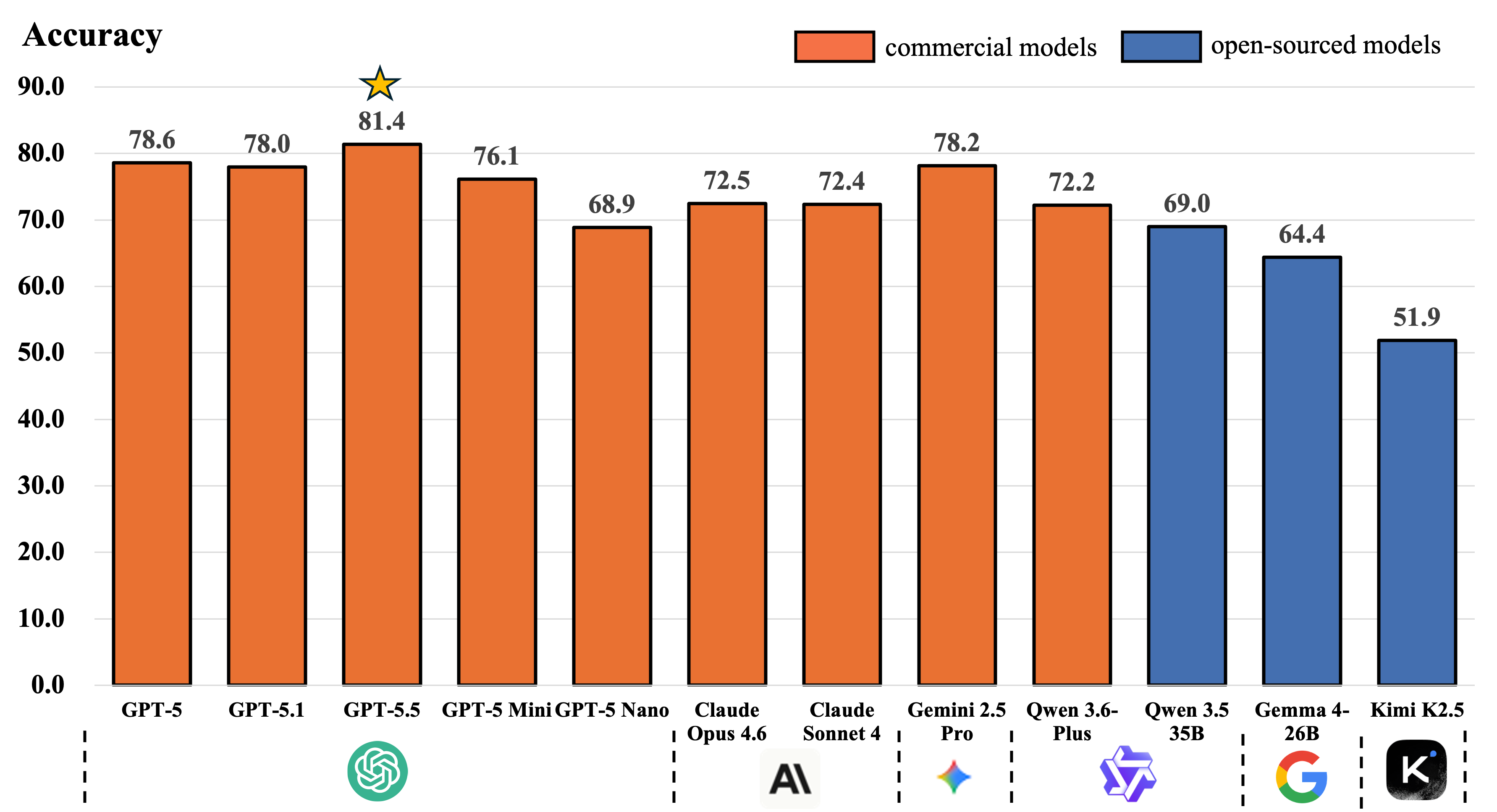}%
  }
  \caption{\textbf{Layer~2 (binary) performance of 12 LLMs.} The gap from near-ceiling MC scores (Tab.~\ref{tab:mc_results}) quantifies the knowledge-application disconnect.}
  \label{fig:overall_accuracy}
\end{figure*}

\paragraph{Multimodal Integrity Check} 
To support multimodal reasoning tasks that require aligned visual and textual evidence across intermediate reasoning steps, we enforce stringent completeness criteria: every retained sample must include both step-by-step instructional text and corresponding images. We implement this through a two-phase verification process. First, an automated filter discards records with missing or malformed metadata fields (as defined in Tab.~\ref{tab:metadata_fields}). Second, human annotators manually inspect the remaining candidates to confirm the presence, alignment, and consistency of multimodal content. This procedure yields 88,723 fully multimodal candidates. We then perform a final curation step to ensure balanced sample distribution and strong category representativeness, with particular attention to the sub-regional diversity of Chinese cuisine, as shown in Tab.~\ref{tab:cuisine_stats}. For each category, we retain examples that satisfy three operationalized criteria: (1)~\textbf{cultural typicality}: the dish is widely recognized as representative of its labeled cuisine (verified by the first author, who has domain expertise in Chinese culinary culture, and cross-checked against cuisine taxonomy references); (2)~\textbf{multimodal consistency}: practice-step images visually correspond to the textual instructions without missing or misaligned steps; and (3)~\textbf{instructional quality}: the recipe contains at least 5 practice steps with clear, non-repetitive photographs. The final procedure yields a curated benchmark of 587 high-quality samples.\footnote{An additional quality audit removed 25 items with incorrect cultural attribution (e.g., ``Mexican bread'', a Chinese bakery category name erroneously classified as Mexican cuisine) and corrected labels for 11 items.} The selection is filtered coverage under fixed, documented criteria rather than an unconstrained hand-picked set. Per-cuisine counts are not forced to be uniform, and thin cuisine classes limit the resolution of per-cuisine estimates.

\paragraph{Dataset Composition}
The 587 curated dishes span 200 Chinese and 387 non-Chinese entries across nine additional countries (Fig.~\ref{fig:language_distribution}, Appendix), totaling \textbf{7,814 images} (avg.\ 12.7 per sample). Chinese dishes constitute 34.1\% of the benchmark and bear fine-grained provincial labels (Tab.~\ref{tab:cuisine_stats}), whereas non-Chinese dishes are labeled by country of origin. This asymmetry reflects two deliberate choices: (1)~Chinese regional cuisines exhibit substantial inter-provincial diversity warranting fine-grained sub-regional study; and (2)~the source platform naturally provides richer Chinese coverage. We report per-language performance throughout (\S\ref{sec:results}) and confirm that Chinese accuracy is comparable to the overall mean, indicating no systematic inflation. The complete metadata schema is summarized in Tab.~\ref{tab:metadata_fields}.

\vspace{-3mm}
\paragraph{Cultural Localization}
For the non-Chinese portion, we restore native dish names and translate procedural instructions according to each dish’s country label. Translation quality was assessed before final curation on 1,236 translated fields from 412 candidate dishes (three fields per dish). Three architecturally diverse models (Gemini-3.0-Pro~\citep{googledeepmind2026gemini3pro}, GPT-5-Mini~\citep{openai2025gpt5}, and DeepSeek-V3~\citep{deepseekai2024deepseekv3}) independently evaluated each field, with a majority vote ($\ge$2/3) required for approval. This procedure attained a macro-average pass rate of \textbf{96.20\%} (Tab.~\ref{tab:translation_quality}); translations that did not pass were manually corrected before release curation. Native speakers additionally audited Hindi, British English, and Vietnamese items against the Chinese sources; details are provided in Appendix~\ref{sec:native_validation}. Detailed error analysis is provided in Appendix~\ref{sec:translation_details}.

Overall, the pipeline's 0.7\% retention rate (587 from 240K raw records) reflects intentionally aggressive filtering to prioritize multimodal completeness and cultural authenticity over scale.

\begin{table}[htbp]
\centering
\resizebox {\columnwidth} {!} {
\begin{tabular}{l|l|m{5cm}}
\toprule
\textbf{Field} & \textbf{Modality} & \textbf{Description} \\ \midrule
\texttt{dish\_name} & Text & Official name of the dish as it appears on the menu. \\ \hline
\texttt{category} & Text & Country of origin with subregion information when available. \\ \hline
\texttt{raw\_material} & Text & Comprehensive list of primary ingredients of the dish. \\ \hline
\texttt{practice\_text} & Text & Ordered, step-by-step instructions for cooking the dish. \\ \hline
\texttt{dish\_image} & Image & High-resolution photograph of the finished dish. \\ \hline
\texttt{practice\_image} & Image & Sequence of photos illustrating each step of the cooking process. \\
\bottomrule
\end{tabular}
}
\caption{\textbf{Metadata schema of \BenchName.}}
\label{tab:metadata_fields}
\end{table}

\subsection{Task Construction}
\label{sec:task_construction}

Leveraging the rich metadata contained in the curated samples (Tab.~\ref{tab:metadata_fields}), we devise 10 distinct evaluation tasks to probe LLMs' understanding and reasoning capabilities across multiple dimensions of culinary culture. These tasks are organized into two tiers:

\paragraph{Recognition tasks} (4 tasks) serve as controlled baselines that test single-step visual or textual matching, which pair dish images or names with categories or practice text. These can typically be solved through direct visual recognition or factual lookup without integrating sequential procedural evidence.

\paragraph{Process-grounded tasks} (6 tasks) require reasoning over multi-step cooking procedures, ingredient--process alignment, or cross-modal procedural matching, which pair practice images with dish images, names, ingredients, or textual instructions. These demand that models track state changes across ordered visual frames and compose intermediate inferences.

Not all tasks in our benchmark require process-grounded reasoning; we retain recognition tasks as controlled baselines to isolate the specific difficulty introduced by procedural understanding. Details of all \TaskNum tasks, including descriptions and prompts, are provided in Tab.~\ref{tab:task_categories} of \S\ref{sec:details_task}. As an example of the most challenging tier, \texttt{practice\_image} $\leftrightarrow$ \texttt{practice\_image} requires LLMs to first identify the dishes based on the practice images, then reason about their culinary origins, and finally verify whether the dishes belong to the same category.

To facilitate automated evaluation, we provide two complementary formats. The \textbf{binary format} frames each task as a yes/no verification question (random baseline: 50\%), while the \textbf{multiple-choice (MC) format} presents four candidate answers per question (random baseline: 25\%), with three carefully constructed distractors sampled from the same cuisine category or task domain to ensure plausibility. Both formats employ balanced data generation: for binary tasks, we produce equal numbers of positive and negative examples; for MC tasks, distractors are drawn from semantically proximate alternatives to preclude elimination by surface cues. All prompts are formulated in native languages to faithfully capture local cultural nuances.

For Chinese dishes, we generate 100 prompts per task, whereas for non-Chinese dishes, the number of prompts per task matches the total count of dishes in their respective category. Consequently, \BenchName contains a total of \SampleNum samples, spanning \LangNum languages across \RegionNum regions. We note that while the 10 tasks draw from a shared pool of 587 base dishes, each task is evaluated independently: no outputs or context are shared across tasks at inference time, and accuracy is computed per task; however, the shared dish pool should still be considered in joint analyses.

\section{Experiments}
\label{sec:results}

\subsection{Setup and Human Baseline}
\label{sec:human_baseline}

We evaluate \BenchName on 12 state-of-the-art LLMs (Fig.~\ref{fig:overall_accuracy}), encompassing both open-weight and proprietary systems. Exploiting the objective nature of our gold answers, we employ fully automated scoring. To contextualize model performance, five annotators with diverse cultural backgrounds independently answer a representative 350-sample subset.

\begin{table}[htbp]
\centering
\small
\begin{tabular}{lcc}
\toprule
\textbf{Annotator} & \textbf{Background} & \textbf{Acc. (\%)} \\
\midrule
Annotator 1 & East Asian cuisine & 87.14 \\
Annotator 2 & European cuisine & 86.86 \\
Annotator 3 & South Asian cuisine & 87.43 \\
Annotator 4 & Chinese regional cuisine & 86.00 \\
Annotator 5 & Global cuisine & 86.86 \\
\midrule
\textbf{Average} & -- & \textbf{86.86} \\
\textbf{Majority Vote} & -- & \textbf{86.86} \\
\bottomrule
\end{tabular}
\caption{\textbf{Human performance on a 350-sample subset of \BenchName.} Pairwise inter-annotator agreement reaches 0.946. The individual-score mean (86.858\%) and majority-vote accuracy (304/350; 86.857\%) independently round to 86.86\%.}
\label{tab:human_baseline}
\end{table}

As shown in Tab.~\ref{tab:human_baseline}, annotators attain 86.00--87.43\% accuracy (avg.\ 86.86\%) with pairwise IAA of 0.946, confirming unambiguous task definitions and reliable gold labels. The $\sim$87\% ceiling establishes that \BenchName is demanding yet well-defined; the best model (GPT-5.5, 81.4\%) still trails humans by $\sim$6 points.

\subsection{The Knowledge-Application Gap}

How well do current LLMs \emph{know} culinary culture? Under the MC format (4,085 items, random baseline 25\%), the answer is: remarkably well. As shown in Tab.~\ref{tab:mc_results}, leading proprietary models attain 94--97\% accuracy, with GPT-5 reaching 96.8\%, Claude Opus 4.6 scoring 96.2\%, and even mid-tier systems like GPT-5~Mini achieving 94.7\%. These scores vastly exceed chance and approach ceiling, spanning all 10 task types and 10 languages. By any conventional metric, this would suggest that current LLMs have internalized substantial culinary cultural knowledge.

Yet this apparent mastery conceals a striking fragility. When the \emph{same} dish--attribute pairings are tested via binary verification like ``Is this dish image correctly paired with [X]?'', accuracy plummets by 10--24 points across the board (Tab.~\ref{tab:mc_results}). The most dramatic case is Claude Opus 4.6, which achieves 96.2\% under MC yet only 72.5\% on binary: a 24-point drop despite near-perfect knowledge confirmation. Even the strongest proprietary model (GPT-5.5: 81.4\%) still trails humans (86.86\%) by $\sim$6 points, while open-weight systems lag considerably further (Qwen3.5-35B: 69.0\%, Gemma4-26B: 64.4\%). This pattern persists across Chinese and non-Chinese subsets (GPT-5: 79.1\% vs.\ 78.3\%; see \S\ref{sec:lang_analysis}).

We term this \textbf{knowledge-application gap}: models that demonstrably \emph{possess} cultural knowledge confirmed by near-perfect MC performance appear unable to \emph{apply} it for independent judgment without candidate contrasts.

\paragraph{Is This Merely a Format Effect?}
A legitimate concern arises: MC and binary verification are structurally distinct tasks (4-way selection vs.\ yes/no judgment; baselines 25\% vs.\ 50\%), so some performance disparity is expected regardless of knowledge. Several observations motivate looking beyond a fixed format penalty. First, the gap varies 2.4$\times$ across models (from 9.8 for Qwen3.5-35B to 23.7 for Claude Opus 4.6). Second, some stronger binary performers show smaller gaps (e.g., GPT-5.5: 14.0 vs.\ Gemma4-26B: 24.0), although this pattern is not uniform. Third, within the GPT family, models with nearly matched MC scores (GPT-5: 96.8\%, GPT-5.1: 96.6\%) also show closely matched binary scores (78.6\% vs.\ 78.0\%). These cross-format patterns are suggestive rather than decisive.

However, the cross-format comparison alone cannot fully disentangle format difficulty from a genuine knowledge-application disconnect. A definitive test requires holding format constant while varying only the \emph{type} of knowledge being probed. So we construct precisely such a test in \S\ref{sec:subregion}, after first validating that the benchmark genuinely demands process reasoning (\S\ref{sec:process_validation}).

\begin{table}[htbp]
\centering
\small
\begin{tabular}{lccr}
\toprule
\textbf{Model} & \textbf{Binary (\%)} & \textbf{MC (\%)} & \textbf{Gap} \\
\midrule
GPT-5.5     & 81.4 & 95.4 & 14.0 \\
GPT-5.1     & 78.0 & 96.6 & 18.6 \\
GPT-5       & 78.6 & 96.8 & 18.2 \\
GPT-5 Nano  & 68.9 & 82.2 & 13.3 \\
GPT-5 Mini  & 76.1 & 94.7 & 18.6 \\
Claude Opus 4.6 & 72.5 & 96.2 & 23.7 \\
Claude Sonnet 4 & 72.4 & 87.1 & 14.7 \\
Gemini 2.5 Pro & 78.2 & 94.6 & 16.4 \\
Qwen3.6-Plus & 72.2 & 95.1 & 22.9 \\
Qwen3.5-35B$^\dagger$ & 69.0 & 78.8 & 9.8 \\
Kimi-K2.5$^\dagger$ & 70.7 & 94.7 & 24.0 \\
Gemma4-26B$^\dagger$  & 64.4 & 88.4 & 24.0 \\
\midrule
Random      & 50.0  & 25.0 & -- \\
\bottomrule
\end{tabular}
\caption{\textbf{Knowledge probe (MC) vs.\ application probe (Binary).} The gap quantifies the knowledge-application disconnect. $^\dagger$Open-source models.}
\label{tab:mc_results}
\end{table}

\subsection{Validating Process-Grounded Reasoning}
\label{sec:process_validation}

Before probing deeper into the knowledge-application gap, we must first establish that our benchmark measures what it claims. If models could solve process-grounded tasks through shallow pattern matching, the gap might simply reflect task difficulty rather than a reasoning deficit. Two analyses corroborate that this is not the case.

\paragraph{Cross-Task Performance.}
The performance heatmap (Fig.~\ref{fig:heatmap}, Appendix) reveals a pronounced hierarchy aligned with our two-tier taxonomy. Recognition tasks yield robust performance (e.g., ``Dish Img $\leftrightarrow$ Practice Text'': GPT-5.1 93.0\%), while more demanding relational and process-grounded tasks deteriorate markedly (``Dish Img $\leftrightarrow$ Dish Img'': 57.8\%, ``Practice Img $\leftrightarrow$ Raw Material'': 58.6\%). This 35-point chasm between the easiest and hardest tasks shows that higher-order relational and process reasoning constitute a distinct and more demanding capability dimension.

\paragraph{Ablation: Removing Sequential Context.}
To verify that sequential visual context drives performance, we conduct an ablation on 500 stratified samples across three models (Qwen3.6-Plus, Claude~4.6, Kimi~K2.5). We compare two conditions: (1)~\textbf{Full}: all practice-step images; (2)~\textbf{Single}: only the first image per step group, removing multi-step process information.

\begin{table}[htbp]
\centering
\small
\begin{tabular}{lccc}
\toprule
\textbf{Model} & \textbf{Full (\%)} & \textbf{Single (\%)} & \textbf{$\Delta$} \\
\midrule
\multicolumn{4}{l}{\textit{Overall}} \\
Claude 4.6    & \textbf{96.4} & 93.0 & $-$3.4 \\
Qwen3.6-Plus  & 95.4 & 88.0 & $-$7.4 \\
Kimi K2.5     & 92.4 & 88.4 & $-$4.0 \\
\midrule
\multicolumn{4}{l}{\textit{Practice-image tasks only}} \\
Claude 4.6    & 97.3 & 91.7 & $-$5.7 \\
Qwen3.6-Plus  & 96.7 & 85.0 & $-$11.7 \\
Kimi K2.5     & 92.3 & 86.3 & $-$6.0 \\
\midrule
\multicolumn{4}{l}{\textit{Non-practice tasks only}} \\
Claude 4.6    & 95.0 & 95.0 & $\pm$0.0 \\
Qwen3.6-Plus  & 93.5 & 92.5 & $-$1.0 \\
Kimi K2.5     & 92.5 & 91.5 & $-$1.0 \\
\bottomrule
\end{tabular}
\caption{\textbf{Ablation study}: Full (all practice-step images) vs.\ Single (one image per step group). Performance degradation confirms process-grounded reasoning dependence on sequential visual information.}
\label{tab:ablation}
\end{table}

As shown in Tab.~\ref{tab:ablation}, all three models exhibit consistent accuracy drops when sequential visual context is reduced: overall accuracy decreases by 3.4--7.4 percentage points. Crucially, this effect is \emph{selective}: practice-image tasks drop by 5.7--11.7pp, while non-practice tasks remain virtually unchanged ($\le$1.0pp). This double dissociation---sequential context matters precisely where process reasoning is required and nowhere else---confirms that our benchmark genuinely demands multi-step visual reasoning rather than shallow pattern matching.

The benchmark passes both tests: it measures what it claims, and it does so through genuine multi-step reasoning. We now turn to the harder question: \emph{what happens when the task demands not merely procedural understanding, but cultural attribution?}

\subsection{Diagnosing Cultural Reasoning Failure}
\label{sec:subregion}

To control response format, we use the same four-way MC interface as standard tasks: given a dish image, models select one of four displayed candidates from a 16-cuisine inventory. The target semantics, label space, and distractors differ; the substantive contrast is standard dish--attribute matching versus regional attribution, the latter requiring seasoning, ingredient, and technique knowledge beyond visual appearance.

We evaluate eight models on 378 four-way MC samples (Tab.~\ref{tab:subregion}), and all score only 38--56\% under the same four-way response interface, compared with up to 97\% on standard tasks. Consider GPT-5.1: 96.6\% on standard MC (Tab.~\ref{tab:mc_results}), yet 48.9\% here, representing a 48 percentage point decline with the response format held fixed. The best-performing models, Kimi~K2.6\footnote{Kimi~K2.6 is the updated release of K2.5 used in Tab.~\ref{tab:mc_results}; we use the newer checkpoint for sub-region evaluation as it became available during our study.} and Claude~4.6, reach only 56.1\%. This within-format comparison constitutes the cleanest evidence for the knowledge-application gap: models can recognize dishes but cannot reliably infer their cultural origins from visual input, even when given the same four-way response scaffolding. Human majority-vote accuracy reaches 69.0\% (Tab.~\ref{tab:human_subregion}; IAA=0.634, reflecting inherent subjectivity), confirming the task is solvable, while the best-performing models still trail humans by 12.9 points.


\begin{table}[htbp]
\centering
\small
\resizebox{\columnwidth}{!}{
\begin{tabular}{lccc}
\toprule
\textbf{Model} & \textbf{Dish Img (\%)} & \textbf{Prac.\ Img (\%)} & \textbf{Overall (\%)} \\
\midrule
Kimi K2.6$^\dagger$   & 59.8 & \textbf{52.4} & \textbf{56.1} \\
Claude 4.6   & \textbf{60.3} & 51.9 & \textbf{56.1} \\
GPT-5.1      & 48.1 & 49.7 & 48.9 \\
Qwen3.6-Plus & 51.3 & 37.0 & 44.2 \\
GPT-5 Mini   & 44.4 & 42.3 & 43.4 \\
Qwen3.5-35B$^\dagger$  & 48.1 & 38.6 & 43.4 \\
GPT-5 Nano   & 46.0 & 39.7 & 42.9 \\
Gemma4-26B$^\dagger$   & 40.7 & 35.4 & 38.1 \\
\midrule
Random (1/4)  & 25.0 & 25.0 & 25.0 \\
\bottomrule
\end{tabular}
}
\caption{\textbf{Layer~3: Sub-region classification.} The large gap from Layer~1 MC performance demonstrates weak visual cultural attribution despite strong standard MC performance. $^\dagger$Open-source models.}
\label{tab:subregion}
\end{table}

\paragraph{Why Do Models Fail?}
\label{sec:diagnosis}
To augment statistical power for the diagnostic analysis below, we expand evaluation to 11 model/configuration conditions spanning 10 unique checkpoints (adding GPT-5, Qwen3.6-Flash, and a high-reasoning GPT-5 variant). Two hypotheses are tenable: (H1)~models attempt cultural reasoning but lack fine-grained precision, or (H2)~models bypass cultural reasoning entirely and rely on visual shortcuts. These hypotheses generate divergent predictions about error patterns, which we adjudicate below.

\paragraph{Confusion Pattern Analysis.}
We group the 16 cuisines into 5 culturally proximate families: Sichuan--Hunan--Guizhou (\begin{CJK}{UTF8}{gbsn}川湘贵\end{CJK}), Cantonese--Fujian--Hakka--Taiwanese (\begin{CJK}{UTF8}{gbsn}粤闽客台\end{CJK}), Shandong--Beijing--Northeastern--Henan (\begin{CJK}{UTF8}{gbsn}鲁北京东北豫\end{CJK}), Zhejiang--Huaiyang--Anhui--Jiangxi (\begin{CJK}{UTF8}{gbsn}浙淮扬徽赣\end{CJK}), and Northwestern (\begin{CJK}{UTF8}{gbsn}西北\end{CJK}). Under H1, errors should cluster within families, as a model with partial knowledge might confuse Cantonese with Hakka, but not with Northeastern. Under H2, errors should disperse uniformly. The data support H2: the intra-family confusion ratio across all 11 model/configuration conditions ranges from 9.8\% to 22.4\%, closely matching the \textbf{random-guessing baseline of 18.1\%} (Tab.~\ref{tab:confusion_diagnosis}). A one-sample binomial test against the 18.1\% baseline yields $p>0.05$ for 9 of 11 conditions; the two exceptions (Gemma4-26B and GPT-5~Mini) deviate \emph{below} chance, showing even less cultural clustering than random, which reinforces rather than undermines H2. When models err on sub-region classification, their mistakes do not exhibit culturally informed structure; instead, they resemble guesses.

\paragraph{Visual Saliency Predicts Accuracy.}
If models rely on visual shortcuts rather than cultural reasoning, we would expect accuracy to track visual distinctiveness rather than cultural complexity. This is precisely what we observe.
Sichuan cuisine (\begin{CJK}{UTF8}{gbsn}川菜\end{CJK}), characterized by red chili oil, Sichuan peppercorns, and an unmistakable color palette, achieves \textbf{78.4\%} average accuracy across all 11 model/configuration conditions.
Anhui dishes (\begin{CJK}{UTF8}{gbsn}徽菜\end{CJK}), such as slow-braised ham and bamboo shoots, often lack distinctive visual markers; accuracy drops to \textbf{18.8\%}, a $4.2\times$ difference from Sichuan cuisine.
Shandong cuisine (\begin{CJK}{UTF8}{gbsn}鲁菜\end{CJK}) is historically prestigious but visually understated, averaging only 29.5\%.
This gradient is remarkably stable: the mean Spearman rank correlation of per-cuisine difficulty across all 55 condition pairs is 0.66, indicating that the models tend to find the same cuisines tractable or intractable. Models perform best when the answer is inscribed on the surface of the image.

\paragraph{Quantity, Not Quality.}
A final question: do stronger models at least \emph{reason differently}, even if imperfectly? They do not. The intra-family confusion ratios of higher-performing models (Kimi~K2.6: 21.7\%, Claude~4.6: 16.9\%) and lower-performing ones (Gemma4-26B: 9.8\%, GPT-5 Mini: 12.6\%) all oscillate around the 18.1\% random baseline. Stronger models simply resolve more visually salient samples correctly. They exhibit superior \emph{pattern discrimination}, not superior \emph{cultural reasoning}. This finding carries a sobering implication: scaling alone appears insufficient to close the knowledge-application gap. The reasoning bridge from perception to cultural knowledge is not reliably activated by visual input.

\paragraph{Text-only Knowledge Probe.}
The diagnostics above demonstrate that models fail at visual cultural attribution, but is the cultural-geographic knowledge itself absent, or merely inaccessible through the visual pathway? To disentangle these possibilities, we remove images entirely and provide \emph{only the dish name} (same 189 samples, identical 4-way options), stripping any cuisine-revealing keywords.\footnote{20/189 dish names contained explicit cuisine keywords. All were cleaned prior to evaluation.} Strikingly, all three models perform \emph{better} without images: Qwen3.6-Plus 68.8\% (+17.5pp), GPT-5 69.8\% (+11.6pp), Kimi~K2.6 67.2\% (+7.4pp; full results in Tab.~\ref{tab:text_only_probe}, Appendix~\ref{sec:cot_combined_appendix}). Models demonstrably \emph{possess} cultural-geographic knowledge (67--70\% from names alone) but cannot mobilize it through visual input. A complementary CoT experiment (Appendix~\ref{sec:cot_combined_appendix}) supports the interpretation that the bottleneck is knowledge activation rather than reasoning depth: CoT yields only +1--2pp, whereas providing dish names alongside images improves by 10--14pp.

\paragraph{Synthesis.}
These diagnostics indicate that models harbor cultural-geographic knowledge in their text representations, but this knowledge becomes inaccessible when routed through visual input. What standard MC evaluations measure as ``cultural knowledge'' is in fact \emph{visual-label discrimination}, not \emph{cultural attribution}. A plausible explanation is that common multimodal objectives emphasize visual--text alignment without sufficient supervision for structured cultural grounding, making the gap difficult to detect under standard MC evaluation. 

\section{Related Work}

\paragraph{Cultural Competence Benchmarks.}
Text-based benchmarks probe factual cultural recall~\citep{DBLP:journals/corr/abs-2205-12247,DBLP:conf/naacl/WangLHJDAC24,myung2024blend} and cultural reasoning~\citep{DBLP:journals/corr/abs-2507-17476,DBLP:conf/acl/HasanHALUSKCA25,DBLP:conf/acl/AroraKCBIC25}. Multimodal evaluations, such as CVQA~\citep{DBLP:conf/nips/RomeroLWGMPOVBJ24}, CaMMT~\citep{DBLP:journals/corr/abs-2505-24456}, and broader suites~\citep{DBLP:conf/iclr/YueSAKNKKSRN25,DBLP:conf/cvpr/VayaniDWASTAHKK25}, assess cultural understanding at scale but remain limited to single-step retrieval or one-hop QA, affording no mechanism to distinguish knowledge \emph{possession} from \emph{application}. Methods for improving cultural awareness, like knowledge augmentation~\citep{DBLP:conf/emnlp/ShiLZZYHPY24}, regional alignment~\citep{DBLP:journals/corr/abs-2504-05154}, targeted training~\citep{DBLP:conf/acl/FengGC0S25}, and red-teaming~\citep{DBLP:journals/corr/abs-2404-06664}, have advanced rapidly, yet verifying whether they yield genuine cultural reasoning requires benchmarks that probe beyond surface-level recognition.

\paragraph{Food-centered Cultural Evaluation.}
Food-centered tasks serve as a natural proxy for cultural competence~\citep{DBLP:conf/naacl/ZhouKLGCCLH25}. FoodieQA~\citep{li2024foodieqa} covers 14 Chinese regional cuisines but lacks procedural context; IndiFoodVQA~\citep{agarwal2024indifoodvqa} introduces multi-hop reasoning but is English-only; WC-VQA~\citep{DBLP:conf/naacl/WinataHIAPWNPOARDPAWMLA25} achieves 30 languages yet offers only two task types with no process metadata. None incorporate step-by-step cooking images, the modality that encodes technique choices and preparation styles distinguishing regional cuisines. \BenchName addresses these lacunae by coupling process-level metadata with sub-regional labels and a three-layer diagnostic protocol that explicitly disentangles knowledge possession from application.

\section{Conclusion}
\vspace{-2pt}
We present \BenchName, a multimodal benchmark (\SampleNum samples, \LangNum languages, \RegionNum regions) evaluating process-grounded culinary reasoning and cultural knowledge. Our evaluation uncovers a \textbf{knowledge-application gap}: leading models attain 94--97\% on standard MC yet reach only 38--56\% on sub-region classification under the same four-way response interface, with diagnostics indicating reliance on visual shortcuts rather than cultural reasoning. Our results suggest that scaling alone is insufficient; future work must explicitly bridge perception and structured cultural knowledge.

\section*{Limitations}
\label{sec:limitations}

\paragraph{Food-specific scope.}
Cuisine, while a rich cultural marker, constitutes but one facet of cultural understanding. Our findings do not necessarily generalize to other domains (history, social customs, arts). Extending the three-layer diagnostic protocol to non-culinary cultural verticals is a natural extension.

\paragraph{Single-platform sourcing.}
All recipes originate from MeishiChina, a Chinese platform. Non-Chinese dishes therefore reflect cross-cultural adaptations rather than natively authored recipes. Future work should incorporate cuisine-native platforms (e.g., Cookpad for Japanese, Giallozafferano for Italian) to complement this perspective. A directional replication on 50 native Italian dishes from Giallozafferano is reported in Appendix~\ref{sec:native_replication}.

\paragraph{Translation quality.}
Multilingual text is LLM-translated and verified by three-model voting (96.2\% macro-average pass rate in the pre-curation audit). However, culturally specific terminology may be transliterated rather than adapted, and LLM judges may share systematic blind spots. We therefore supplement the tribunal with targeted native-speaker validation (Appendix~\ref{sec:native_validation}), while broader native review remains future work.

\paragraph{Data reuse across tasks.}
All 10 tasks draw from the same 587-dish pool. Although each task is evaluated independently, users conducting joint analyses should be aware of potential memorization effects across task boundaries. Scaling to separate dish pools per task would strengthen independence guarantees.

\paragraph{Objective-only evaluation.}
Binary and MC formats enable automated scoring but exclude items requiring subjective interpretation or culturally contingent appropriateness, which are competencies central to authentic cross-cultural communication. Incorporating open-ended generation with human or LLM-as-judge evaluation would complement our diagnostic protocol.

\section*{Ethical Considerations}

\BenchName is built from publicly accessible, user-contributed recipes on
MeishiChina, collected solely for research evaluation; the released metadata
contains no user names, account identifiers, or other personal information.
Human evaluation was performed voluntarily by informed colleagues who are not
authors of this paper, and only anonymized, aggregate results are reported.
Risks of cultural misrepresentation arising from single-platform sourcing and
translation are discussed in the Limitations section. AI assistants
(Grammarly, ChatGPT, and Claude) were used only for grammar checking and
rephrasing; all research ideas, experiments, and analyses are the authors'
own. Code, data, and reproducibility materials are available.

\section*{Acknowledgments}
We thank the anonymous reviewers and the area chair for their valuable feedback and constructive suggestions. This research was jointly supported by Alibaba Group and the Excellence Funds of Universität Hamburg.

\bibliography{custom}

\appendix

\section{Scope of the Diagnostic Evidence}
\label{sec:scope_appendix}

Our confusion and saliency analyses are grounded in Chinese regional cuisines, which is the only subset with fine-grained sub-regional labels. However, three observations suggest the diagnosed mechanism generalizes beyond this specific cultural context. First, the failure is architecture-agnostic: it manifests across 11 model/configuration conditions spanning 10 unique checkpoints, multiple model families, and both open-source and commercial systems. Second, the distinction between visual discrimination and cultural attribution is not specific to Chinese food; it applies equally to domains where surface appearance diverges from categorical membership. Third, our Layer~1--2 results show that the MC-to-binary gap is consistent across all 10 languages (12--26pp; Appendix~\ref{sec:additional_analyses}), indicating that the knowledge-application disconnect is cross-cultural even where we lack sub-regional labels to perform the full diagnostic. We view the Chinese sub-region analysis as a detailed case study of a general phenomenon. A directional pilot on 50 native Italian dishes reproduces the recognition--region gap (Appendix~\ref{sec:native_replication}), but broader validation requires scaling verified sub-regional labels, including Italian regional and Indian state-level cuisines. Future expansion can combine native platforms, LLM-assisted pre-filtering with human adjudication, and audited community contributions.

\section{Native-Data Replication beyond Chinese Cuisine}
\label{sec:native_replication}

To test whether the knowledge-application gap is specific to cross-cultural adaptation, translation, or the source platform, we collect 50 native Italian dishes from Giallozafferano\footnote{\url{https://www.giallozafferano.com/}}, using region labels from its ``ricette regionali'' taxonomy. We rerun dish recognition and region attribution as four-way MC tasks (25\% random baseline) on four models.

\begin{table}[H]
\centering
\small
\resizebox{\columnwidth}{!}{
\begin{tabular}{lccc}
\toprule
\textbf{Model} & \textbf{Recognition (\%)} & \textbf{Region (\%)} & \textbf{Gap (pp)} \\
\midrule
GPT-5.1          & 96.0 & \textbf{80.0} & 16.0 \\
Kimi K2.6$^\dagger$      & 96.0 & 68.0 & 28.0 \\
Qwen3.6-Plus     & 92.0 & 56.0 & 36.0 \\
Qwen3.5-35B$^\dagger$    & \textbf{98.0} & 60.0 & 38.0 \\
\bottomrule
\end{tabular}
}
\caption{Native Italian replication ($N=50$, four-way MC). The recognition--region gap appears across all four models. $^\dagger$Open-source models.}
\label{tab:native_replication}
\end{table}

The gap reproduces in direction for every model (16--38pp), with recognition at or above 92\%. Because this pilot covers only 50 dishes, it provides directional cross-platform evidence rather than a comprehensive estimate of Italian regional-cuisine performance.

\section{Confusion Diagnosis}
\label{sec:confusion_appendix}

\begin{table}[H]
\centering
\small
\resizebox{\columnwidth}{!}{
\begin{tabular}{lcccc}
\toprule
\textbf{Model} & \textbf{Acc.\ (\%)} & \textbf{Errors} & \textbf{Intra-fam.\ (\%)} & \textbf{$\Delta$ vs.\ rand.} \\
\midrule
GPT-5$^\ddagger$  & 57.1 & 162 & 17.9 & $-$0.2 \\
Claude 4.6        & 56.1 & 166 & 16.9 & $-$1.2 \\
Kimi K2.6$^\dagger$ & 56.1 & 166 & 21.7 & +3.6 \\
GPT-5             & 53.2 & 177 & 14.7 & $-$3.4 \\
GPT-5.1           & 48.9 & 193 & 18.7 & +0.6 \\
Qwen3.6-Flash     & 46.6 & 202 & 16.3 & $-$1.8 \\
Qwen3.6-Plus      & 44.2 & 211 & 20.9 & +2.8 \\
GPT-5 Mini        & 43.4 & 214 & 12.6 & $-$5.5 \\
Qwen3.5-35B$^\dagger$ & 43.4 & 214 & 22.4 & +4.3 \\
GPT-5 Nano        & 42.9 & 216 & 15.7 & $-$2.4 \\
Gemma4-26B$^\dagger$ & 38.1 & 234 & 9.8 & $-$8.3 \\
\midrule
Random baseline   & 25.0 & --- & 18.1 & 0.0 \\
\bottomrule
\end{tabular}
}
\caption{Diagnosing the cultural reasoning gap across 11 model/configuration conditions (10 unique checkpoints). \textbf{Intra-fam.}: percentage of errors where the predicted cuisine belongs to the same cultural family as the ground truth. \textbf{$\Delta$ vs.\ rand.}: deviation from the 18.1\% random baseline. No condition shows meaningful above-chance cultural clustering. $^\dagger$Open-source. $^\ddagger$High reasoning effort.}
\label{tab:confusion_diagnosis}
\end{table}

\section{Additional Analyses}
\label{sec:additional_analyses}

\begin{figure*}
    \resizebox{\textwidth}{!}{%
      \includegraphics{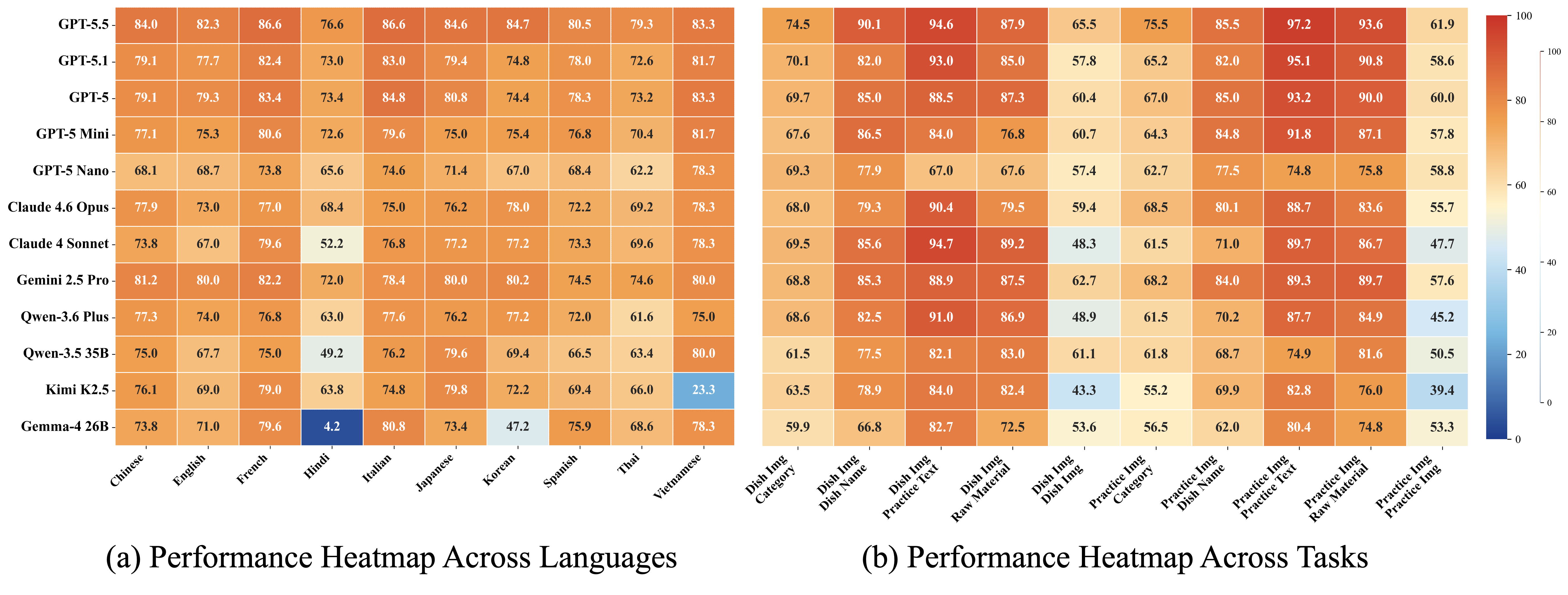}%
    }
    \captionof{figure}{Performance heatmap of 12 LLMs on \BenchName across different languages and tasks.}
    \label{fig:heatmap}
\end{figure*}

\paragraph{Cross-Lingual Consistency.}
\label{sec:lang_analysis}
Figure~\ref{fig:heatmap} (a) shows performance variation across languages, where high-resource European languages lead (French 83.4\%, Italian 84.8\%) while low-resource Asian languages trail (Korean 73.2\%, Thai 73.2\%). Crucially, the MC-to-binary gap persists across all ten languages (range: 12--26pp), confirming that the knowledge-application disconnect is not an artifact of any particular linguistic or cultural subset.

\paragraph{Response Bias.}
Ground-truth labels are approximately uniform across A/B/C/D. Most models preserve this balance ($\chi^2$ test, $p>0.05$), though GPT-5~Nano shows a significant preference for option~A (29.1\%, $p<0.0001$). In binary tasks, both annotators and models exhibit a rejection bias (58--65\% ``no'' responses despite balanced ground truth), indicating genuine task difficulty rather than annotation artifacts.

\paragraph{Disjoint-Subset and Bootstrap Analysis.}
\label{sec:subset_bootstrap}
We split the 177 dishes shared by the Layer~1 and Layer~3 MC tasks into five disjoint random subsets and recompute the within-format gap for the six models evaluated on both layers. The gap remains positive in every subset for every model (no subset gap below 0.30), with a standard deviation across subsets of at most 0.098. Across 1,000 bootstrap samples, the proportion with a positive gap is 1.000 for all six models, showing that the aggregate gap is not carried by a small set of dishes.

\paragraph{Reasoning Chain Verification.}
\label{sec:cot}
On Layer~2 process-reasoning tasks (where models perform well), do they actually engage in structured multi-step reasoning, or succeed through shortcuts? We collect reasoning chains from Claude~4.6 and Qwen3.6-Plus on 30 practice-image samples. Both exhibit structured patterns: visual observation (83--100\%), process-level reasoning (87--93\%), and ingredient--procedure cross-referencing (50--93\%). This confirms that the benchmark elicits genuine multi-step reasoning on process tasks, making the Layer~3 cultural reasoning failure all the more striking: models \emph{can} reason over procedures, but cannot bridge from procedural understanding to cultural attribution.

\section{Supplementary Tables}
\label{sec:supplementary_tables}

\begin{table}[H]
\centering
\small
\resizebox{\columnwidth}{!}{
\begin{tabular}{l|c|c}
\toprule
\textbf{Language} & \textbf{Pass Rate (\%)} & \textbf{Passed/Total} \\
\midrule
English / Spanish / French & 94.4 / 96.1 / 96.7 & 449/468 \\
Hindi / Italian / Thai & 94.0 / 95.3 / 93.3 & 424/450 \\
Japanese / Korean / Viet. & 98.7 / 97.3 / 100.0 & 312/318 \\
\midrule
\textbf{Average} & \textbf{96.20} & \textbf{-} \\
\bottomrule
\end{tabular}
}
\caption{Pre-curation translation audit on 412 candidate non-Chinese dishes (1,236 fields) via an LLM voting tribunal (Gemini-3.0-Pro, GPT-5-Mini, DeepSeek-V3). A majority vote ($\ge$2/3) is required for approval; 96.20\% is the macro-average across languages.}
\label{tab:translation_quality}
\end{table}

\begin{table}[H]
\centering
\small
\resizebox{\columnwidth}{!}{
\begin{tabular}{l c l c}
    \toprule
    \textbf{Regional Cuisine} & \textbf{Count} & \textbf{Regional Cuisine} & \textbf{Count} \\
    \midrule
    Lu Cuisine (Shandong)     & 32             & Guizhou Cuisine       & 6              \\
    Yue Cuisine (Cantonese)   & 28             & Hakka Cuisine           & 5              \\
    Chuan Cuisine (Sichuan)   & 21             & Taiwanese Cuisine      & 4              \\
    Xiang Cuisine (Hunan)     & 19             & Yu Cuisine (Henan)     & 4              \\
    Zhe Cuisine (Zhejiang)    & 15             & Gan Cuisine (Jiangxi)   & 4              \\
    Beijing Cuisine           & 11             & Yunnan Cuisine          & 3              \\
    Northeastern Cuisine      & 10             & Hong Kong Cuisine      & 2              \\
    Northwestern Cuisine      & 8              & Su Cuisine (Jiangsu)     & 2              \\
    Min Cuisine (Fujian)      & 8              & Jin Cuisine (Shanxi)    & 2              \\
    Huaiyang Cuisine          & 7              & Xinjiang Cuisine      & 1              \\
    Hui Cuisine (Anhui)       & 7              & E Cuisine (Hubei)      & 1              \\
    \bottomrule
\end{tabular}
}
\caption{Distribution of dishes by regional Chinese cuisine in \BenchName.}
\label{tab:cuisine_stats}
\end{table}

\begin{table}[H]
\centering
\small
\resizebox{\columnwidth}{!}{
\begin{tabular}{lcc}
\toprule
\textbf{Annotator} & \textbf{Background} & \textbf{Acc.\ (\%)} \\
\midrule
Annotator 1 & Sichuan/Hunan cuisine & 65 \\
Annotator 2 & Cantonese/Fujian cuisine & 55 \\
Annotator 3 & Shandong/Northeastern cuisine & 63 \\
Annotator 4 & Zhejiang/Huaiyang cuisine & 60 \\
Annotator 5 & Generalist food critic & 63 \\
\midrule
\textbf{Average} & -- & \textbf{61.2} \\
\textbf{Majority Vote} & -- & \textbf{69.0} \\
\bottomrule
\end{tabular}
}
\caption{Human performance on sub-region classification (100 samples, 16 Chinese regional cuisines, 4-way MC). Pairwise inter-annotator agreement: 0.634.}
\label{tab:human_subregion}
\end{table}

\section{Sample Distribution}
\label{sec:distribution_appendix}

\begin{figure}[H]
  \centering
  \resizebox{\columnwidth}{!}{%
  \includegraphics{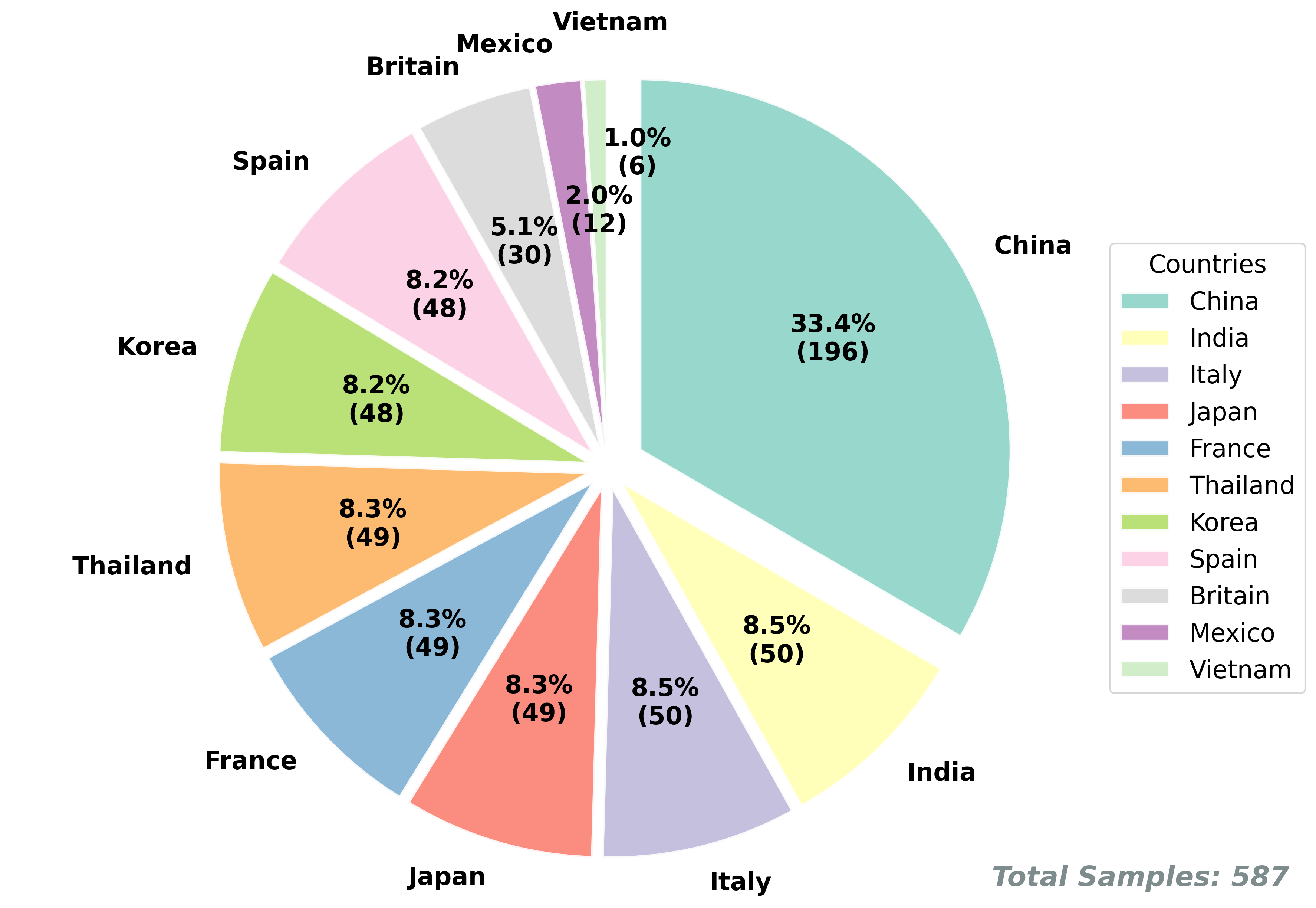}%
  }
  \caption{Distribution of the curated menu samples across 10 countries.}
  \label{fig:language_distribution}
\end{figure}

\section{Details of Tasks}
\label{sec:details_task}
Details of the 10 tasks in CulturalMenuBench are listed in Tab.~\ref{tab:task_categories}. For each task, two metadata fields were selected for matching, and we ensured that each task included image metadata to create complex multimodal reasoning tasks. To ensure objective evaluation, we constrained the model to make binary judgments (yes or no) in the prompts for each task. Additionally, for cuisines from different countries, we used the corresponding national language in the prompts (not shown in the table), which effectively evaluates the model’s multilingual understanding capabilities.
\begin{table*}[t]
\centering
\small
\renewcommand{\arraystretch}{1.15}
\begin{tabularx}{\textwidth}{|X|}
\hline
\textbf{Task:} \texttt{dish\_image} $\leftrightarrow$ \texttt{dish\_name}\\
Description: Given a dish image and a candidate dish name, decide whether the name correctly identifies the food shown.\\
Prompt: ``Is this image \textit{[dish\_name]}? \textit{[dish\_image]}''\\ \hline

\textbf{Task:} \texttt{dish\_image} $\leftrightarrow$ \texttt{category}\\
Description: Given a dish image and a cuisine label, judge whether the dish belongs to that cuisine.\\
Prompt: ``Is this image \textit{[category]} cuisine? \textit{[dish\_image]}''\\ \hline

\textbf{Task:} \texttt{dish\_image} $\leftrightarrow$ \texttt{raw\_material}\\
Description: Given a dish image and a list of main ingredients, verify whether the ingredients match the pictured dish.\\
Prompt: ``Is the main ingredient of this dish \textit{[raw\_material]}? \textit{[dish\_image]}''\\ \hline

\textbf{Task:} \texttt{dish\_image} $\leftrightarrow$ \texttt{practice\_text}\\
Description: Given a dish image and a cooking-method description, decide whether the method matches the dish.\\
Prompt: ``The cooking method for this dish is: \textit{[practice\_text]}. Is this description correct? \textit{[dish\_image]}''\\ \hline

\textbf{Task:} \texttt{practice\_image} $\leftrightarrow$ \texttt{dish\_name}\\
Description: Given several cooking-process images and a dish name, determine whether the images depict the preparation of that dish.\\
Prompt: ``Do these cooking process images show the preparation of \textit{[dish\_name]}? \textit{[practice\_image]}''\\ \hline

\textbf{Task:} \texttt{practice\_image} $\leftrightarrow$ \texttt{category}\\
Description: Given several cooking-process images and a cuisine label, judge whether the technique belongs to that cuisine.\\
Prompt: ``Do these cooking process images show the preparation of \textit{[category]} cuisine? \textit{[practice\_image]}''\\ \hline

\textbf{Task:} \texttt{practice\_image} $\leftrightarrow$ \texttt{raw\_material}\\
Description: Given several cooking-process images and a list of main ingredients, verify whether the ingredients fit the depicted dish.\\
Prompt: ``Do these cooking process images show a dish whose main ingredient is \textit{[raw\_material]}? \textit{[practice\_image]}''\\ \hline

\textbf{Task:} \texttt{practice\_image} $\leftrightarrow$ \texttt{practice\_text}\\
Description: Given several cooking-process images and a cooking-method description, decide whether they correspond.\\
Prompt: ``The cooking method shown in these images is: \textit{[practice\_text]}. Is this description correct? \textit{[practice\_image]}''\\ \hline

\textbf{Task:} \texttt{dish\_image} $\leftrightarrow$ \texttt{dish\_image}\\
Description: Given two dish images, determine whether they belong to the same cuisine category.\\
Prompt: ``Do the following two dish images belong to the same cuisine category? \texttt{dish\_image1} \texttt{dish\_image2}''\\ \hline

\textbf{Task:} \texttt{practice\_image} $\leftrightarrow$ \texttt{practice\_image}\\
Description: Given two sets of cooking-process images, assess whether they represent dishes from the same cuisine category.\\
Prompt: ``Do the following two groups of preparation step images belong to the same cuisine category? \texttt{practice\_image1} \texttt{practice\_image2}''\\ \hline
\end{tabularx}
\caption{Task categories and their corresponding prompt templates. Each task is expressed in three consecutive lines: task name, description, and prompt. All tasks require a binary (yes/no) answer expressed in the target language.}
\label{tab:task_categories}
\end{table*}

\section{Multiple-Choice Format Details}
\label{sec:mc_details}
To complement the binary evaluation, we construct a multiple-choice (MC) variant of \BenchName comprising 4,085 items across 14 directed relations. For each item, three plausible distractors are sampled from the same cuisine category or metadata domain as the correct answer, ensuring that surface-level elimination strategies are ineffective. The MC format reduces the random baseline from 50\% (binary) to 25\%, providing a more discriminating evaluation of model capabilities. The per-task MC accuracy for complete models is reported in Tab.~\ref{tab:mc_per_task}.

\begin{table}[H]
\centering
\small
\resizebox{\columnwidth}{!}{
\begin{tabular}{lccc}
\toprule
\textbf{Task} & \textbf{GPT-5 Mini} & \textbf{Claude Son.\ 4} & \textbf{GPT-5 Nano} \\
\midrule
DI $\leftrightarrow$ DN & 92 & 86 & 88 \\
DI $\leftrightarrow$ PI & 95 & 82 & 75 \\
DI $\leftrightarrow$ PT & 97 & 92 & 86 \\
DI $\leftrightarrow$ RM & 94 & 86 & 85 \\
DN $\leftrightarrow$ DI & 93 & 86 & 84 \\
DN $\leftrightarrow$ PI & 90 & 82 & 78 \\
PI $\leftrightarrow$ DI & 97 & 82 & 54 \\
PI $\leftrightarrow$ DN & 92 & 84 & 82 \\
PI $\leftrightarrow$ PT & 99 & 94 & 88 \\
PI $\leftrightarrow$ RM & 96 & 88 & 91 \\
PT $\leftrightarrow$ DI & 96 & 92 & 84 \\
PT $\leftrightarrow$ PI & 98 & 93 & 81 \\
RM $\leftrightarrow$ DI & 94 & 85 & 81 \\
RM $\leftrightarrow$ PI & 96 & 83 & 78 \\
\midrule
Overall & 94.7 & 87.1 & 82.2 \\
\bottomrule
\end{tabular}
}
\caption{Per-task MC accuracy (\%). DI = Dish Image, DN = Dish Name, PI = Practice Image, PT = Practice Text, RM = Raw Material. Practice-image tasks (PI $\leftrightarrow$ DI) show the largest gap between models, with GPT-5 Nano dropping to 54\%.}
\label{tab:mc_per_task}
\end{table}

\section{Translation Quality Details}
\label{sec:translation_details}

The LLM voting tribunal exhibits meaningful inter-judge disagreement: only 84.0\% of items receive unanimous 3/3 votes, with the remaining 16\% showing split decisions, confirming that judges are not rubber-stamping translations. Error analysis of the rejected items reveals that information omission (46\%) is the dominant failure mode (e.g., omitting ``Korean-style'' from a dish name), followed by terminology errors (6\%) and unnaturalness (6\%). Failure rates vary across both field and cuisine. Among fields, dish names are the most difficult to translate, with a 90.0\% pass rate, compared with 98.5\% for procedures and 99.3\% for ingredient lists. Across cuisines, Indian and British cuisines have lower pass rates for dish names, at 80\% and 83\%, respectively, suggesting greater cross-cultural variation in naming conventions.

\paragraph{Native-Speaker Validation.}
\label{sec:native_validation}
Native speakers who read Chinese audited 18 items each in Hindi, British English, and Vietnamese against the Chinese sources, covering dish names, ingredients, and cooking steps. The annotations were pre-drafted with LLM assistance and then reviewed and signed off by the native speakers. The audit found no outright mistranslations and 2--3 under-specification cases per language, such as a specific dish receiving a more generic name. It also identified a duplicated source string and residual Chinese punctuation. These cases are disambiguated or corrected in the release.

\section{Error Analysis by Failure Type}
\label{sec:error_analysis_appendix}

We analyze errors from a high-performing model (GPT-5.1, 3.4\% error rate) and a lower-performing full-coverage model (GPT-5~Nano, 17.8\% error rate), classifying them into three categories.

\paragraph{Visual Confusion} accounts for 49.6\% of GPT-5.1 errors but only 20.7\% of GPT-5~Nano errors. These errors arise from visually similar dishes (e.g., confusing two curry-based soups or pasta variants). Indian cuisine is disproportionately affected due to its curry-heavy visual homogeneity, accounting for 24.7\% of shared hard cases despite being only 11.8\% of the data.

\paragraph{Process Reasoning Failure} is the dominant error mode for GPT-5~Nano (51.4\% of its errors) but only 22.6\% for GPT-5.1. The hardest task, \texttt{practice\_image$\to$dish\_image}, yields a 46.0\% error rate for Nano vs.\ 1.0\% for GPT-5.1, which suggests that reasoning about cooking procedures scales steeply with model capacity.

\paragraph{Ingredient Recognition Failure} constitutes $\sim$28\% of errors for both models, though the absolute count in Nano is 6.3$\times$ higher, indicating that ingredient identification difficulty scales proportionally with overall capability.

\paragraph{Language and Cuisine Effects.} Vietnamese (57 samples) has the highest error rate for both models: 13.7\% for GPT-5.1 and 36.8\% for Nano. Together, Vietnamese and Hindi account for 28.7\% of GPT-5.1 errors despite representing only 13.4\% of the data, revealing compound difficulty from low-resource languages paired with visually homogeneous cuisines. Of GPT-5.1's 115 errors, 85 (74\%) are shared with Nano, indicating that residual errors cluster on genuinely ambiguous items.

\section{Experimental Setup Details}
\label{sec:exp_setup_appendix}

\paragraph{Model Configurations.}
Table~\ref{tab:model_configs} summarizes the 12 models evaluated in Layers~1--2 and the 11 model/configuration conditions (10 unique checkpoints) used for Layer~3 sub-region diagnosis. Model provenance is documented by the corresponding system cards and technical reports for GPT-5/5.1/5.5~\citep{openai2025gpt5,openai2025gpt51,openai2026gpt55}, Claude~4/4.6~\citep{anthropic2025claude4,anthropic2026claudeopus46}, Gemini~2.5~\citep{geminiteam2025gemini25}, Kimi~K2.5~\citep{kimiteam2026k25}, and Gemma~4~\citep{gemmateam2026gemma4}. All API-based models are accessed via official endpoints; the exact provider documentation is available for Kimi~K2.6\footnote{Official API documentation: \href{https://platform.moonshot.ai/docs/guide/use-kimi-api}{Moonshot AI platform}.} and Qwen3.6-Plus/Flash.\footnote{Official model pages: \href{https://help.aliyun.com/zh/model-studio/qwen3-6-plus}{Qwen3.6-Plus} and \href{https://help.aliyun.com/zh/model-studio/qwen3-6-flash}{Qwen3.6-Flash}.} The Qwen3.5-35B checkpoint follows its official model card.\footnote{Official model card: \href{https://huggingface.co/Qwen/Qwen3.5-35B-A3B}{Qwen3.5-35B-A3B}.} Open-weight models are served locally via vLLM~v0.19.1~\citep{kwon2023vllm} with tensor parallelism across 2 GPUs.

\begin{table}[H]
\centering
\small
\resizebox{\columnwidth}{!}{
\begin{tabular}{lll}
\toprule
\textbf{Model} & \textbf{Identifier} & \textbf{Access} \\
\midrule
\multicolumn{3}{l}{\textit{Proprietary (API)}} \\
GPT-5.5 & gpt-5.5-2025-11-20 & OpenAI \\
GPT-5 & gpt-5-2025-08-07 & OpenAI \\
GPT-5.1 & gpt-5.1-2025-11-13 & OpenAI \\
GPT-5 Mini & gpt-5-mini-2025-08-07 & OpenAI \\
GPT-5 Nano & gpt-5-nano-2025-08-07 & OpenAI \\
Gemini 2.5 Pro & gemini-2.5-pro & Google AI \\
Claude Opus 4.6 & claude-opus-4-6 & Anthropic \\
Claude Sonnet 4 & claude-sonnet-4 & Anthropic \\
Kimi K2.5/K2.6 & kimi-k2.5 / kimi-k2.6 & Moonshot AI \\
Qwen3.6-Plus & qwen3.6-plus-2026-04-02 & Alibaba \\
Qwen3.6-Flash & qwen3.6-flash-2026-04-02 & Alibaba \\
\midrule
\multicolumn{3}{l}{\textit{Open-weight (local inference)}} \\
Qwen3.5-35B & Qwen/Qwen3.5-35B-A3B & vLLM, TP=2 \\
Gemma4-26B & gemma-4-26B-A4B-it & vLLM, TP=2 \\
\bottomrule
\end{tabular}
}
\caption{Model configurations. All OpenAI models use the GlobalStandard variant. Kimi~K2.6 is used for sub-region evaluation (Layer~3); K2.5 for Layers~1--2. Open-weight models are served with BF16 precision and max sequence length of 8,192 tokens.}
\label{tab:model_configs}
\end{table}

\paragraph{Inference Parameters.}
For Layer~1--2 evaluation, all models use greedy decoding (temperature=0) with max output tokens of 64 (binary and MC tasks require only short outputs). For the sub-region diagnosis (Layer~3), GPT-family models use \texttt{reasoning\_effort=low} to reduce latency; Claude Opus 4.6 uses extended thinking with a budget of 1,024 tokens. The text-only probe uses temperature=0 with max 64 tokens.

\paragraph{Prompt Templates.}
We provide the exact prompts used for each experimental condition:

\begin{itemize}
\item \textbf{Standard sub-region MC} (image input):
\begin{quote}
\small\ttfamily
[dish image]\\
\begin{CJK}{UTF8}{gbsn}这张菜品图片属于以下哪个菜系？\end{CJK}\\
A. \begin{CJK}{UTF8}{gbsn}贵州菜\end{CJK} \quad B. \begin{CJK}{UTF8}{gbsn}浙菜\end{CJK} \quad C. \begin{CJK}{UTF8}{gbsn}客家菜\end{CJK} \quad D. \begin{CJK}{UTF8}{gbsn}川菜\end{CJK}\\
\begin{CJK}{UTF8}{gbsn}请仅输出A、B、C、D选项的字母，不要输出任何其他内容。\end{CJK}
\end{quote}

\item \textbf{Text-only probe} (no image, dish name input):
\begin{quote}
\small\ttfamily
\begin{CJK}{UTF8}{gbsn}根据菜名判断，"\{dish\_name\}" 属于以下哪个菜系？\end{CJK}\\
A. ... \quad B. ... \quad C. ... \quad D. ...\\
\begin{CJK}{UTF8}{gbsn}请仅输出A、B、C、D选项的字母，不要输出任何其他内容。\end{CJK}
\end{quote}
Dish names are cleaned to remove cuisine-revealing keywords (e.g., \begin{CJK}{UTF8}{gbsn}``四川小吃红油抄手''\end{CJK}$\to$\begin{CJK}{UTF8}{gbsn}``红油抄手''\end{CJK}). 20/189 names required cleaning.

\item \textbf{CoT prompting} (image + reasoning instruction):
\begin{quote}
\small\ttfamily
[dish image]\\
\begin{CJK}{UTF8}{gbsn}请仔细观察这张菜品图片，按以下步骤推理：\end{CJK}\\
\begin{CJK}{UTF8}{gbsn}1. 首先识别这道菜是什么\end{CJK}\\
\begin{CJK}{UTF8}{gbsn}2. 分析其食材、调味料和烹饪手法的特征\end{CJK}\\
\begin{CJK}{UTF8}{gbsn}3. 根据这些特征判断它最可能属于哪个菜系\end{CJK}\\
A. ... \quad B. ... \quad C. ... \quad D. ...\\
\begin{CJK}{UTF8}{gbsn}请先给出你的推理过程，最后在新的一行输出答案字母。\end{CJK}
\end{quote}

\item \textbf{Text+Image combined} (dish name + image):
\begin{quote}
\small\ttfamily
[dish image]\\
\begin{CJK}{UTF8}{gbsn}这道菜叫做「\{dish\_name\}」。结合菜名和图片，判断它属于以下哪个菜系？\end{CJK}\\
A. ... \quad B. ... \quad C. ... \quad D. ...\\
\begin{CJK}{UTF8}{gbsn}请仅输出A、B、C、D选项的字母，不要输出任何其他内容。\end{CJK}
\end{quote}
\end{itemize}

\begin{figure*}
  \resizebox{\linewidth}{!}{%
  \includegraphics{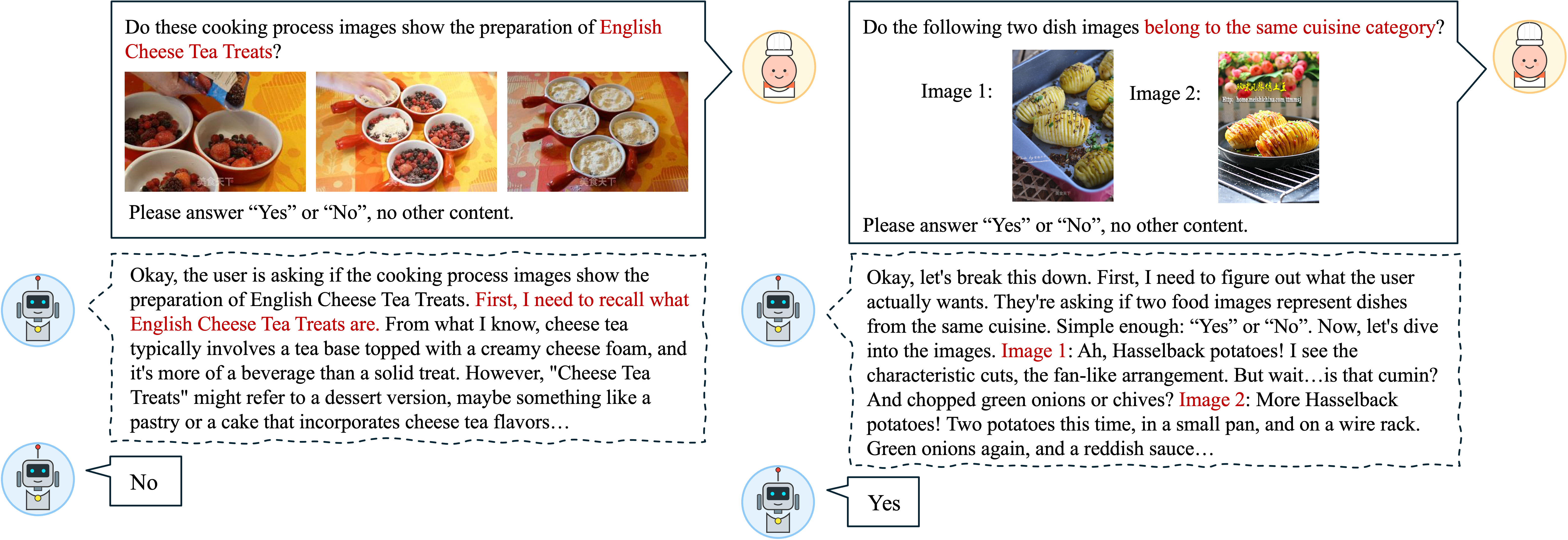}%
  }
  \captionof{figure}{Two representative examples from \BenchName. Task~1 requires cross-cultural semantic disambiguation; Task~2 requires abstracting away surface-level plating differences to recognize categorical equivalence. Both demand multi-step reasoning integrating visual, procedural, and cultural knowledge.}
  \label{fig:case}
\end{figure*}

\paragraph{Distractor Construction.}
For all 4-way MC tasks (Layers~1--3), distractors are sampled from the same cultural family when possible. For sub-region classification, each item's three distractors are drawn from the pool of 16 regional cuisines, with at least one from the same cultural family (e.g., for a Sichuan dish: one Hunan/Guizhou option plus two from other families). This ensures that surface-level elimination is insufficient.

\paragraph{Evaluation Protocol.}
Answer extraction uses regex matching for the first (standard tasks) or last (CoT) occurrence of [A-D] in model output. Samples with empty responses or unparseable outputs are retried up to 3--5 times with exponential backoff. The concurrency is set to 4--10 parallel requests per model depending on API rate limits.

\section{CoT and Text-Anchor Ablation}
\label{sec:cot_combined_appendix}


\paragraph{CoT Prompting.} On the same 177 dish-image samples, we instruct models to reason step-by-step before answering: (1)~identify the dish, (2)~analyze ingredients, seasonings, and cooking techniques, (3)~determine the most likely regional cuisine. All models employ their strongest available reasoning configuration (GPT-5: built-in reasoning tokens; Qwen3.6-Plus: extended thinking enabled; Gemma4-26B: prompt-level CoT).

\paragraph{Text+Image Combined.} We provide the dish name alongside the image (e.g., ``Based on the name `\begin{CJK}{UTF8}{gbsn}宫保鸡丁\end{CJK}' and image, which cuisine does it belong to?''), testing whether textual anchoring can bridge the visual-to-cultural reasoning gap.

\begin{table}[H]
\centering
\small
\begin{tabular}{lccc}
\toprule
\textbf{Model} & \textbf{Baseline} & \textbf{+CoT} & \textbf{+Name} \\
\midrule
GPT-5        & 57.6 & 59.3 \small{(+1.7)} & \textbf{67.8} \small{(+10.2)} \\
Qwen3.6-Plus & 58.3 & 59.3 \small{(+1.0)} & \textbf{67.8} \small{(+9.5)} \\
Gemma4-26B   & 43.5 & 45.2 \small{(+1.7)} & \textbf{57.1} \small{(+13.6)} \\
\bottomrule
\end{tabular}
\caption{CoT prompting vs.\ text-anchor ablation on sub-region classification. CoT yields marginal improvement, whereas providing the dish name boosts accuracy by 10--14pp, supporting the interpretation that the bottleneck is knowledge activation rather than reasoning depth. Notably, even the strongest text-anchored result (67.8\%) only \emph{matches} text-only performance rather than exceeding it, suggesting that visual input contributes little additive cultural signal beyond what the dish name alone provides.}
\label{tab:cot_combined}
\end{table}

\begin{table}[H]
\centering
\small
\begin{tabular}{lccc}
\toprule
\textbf{Model} & \textbf{Image (\%)} & \textbf{Text-only (\%)} & \textbf{$\Delta$} \\
\midrule
Qwen3.6-Plus & 51.3 & \textbf{68.8} & +17.5 \\
Kimi K2.6    & 59.8 & \textbf{67.2} & +7.4 \\
GPT-5        & 58.2 & \textbf{69.8} & +11.6 \\
\bottomrule
\end{tabular}
\caption{Text-only vs.\ image-based sub-region classification (same 189 samples, same 4-way MC options). Removing images \emph{improves} accuracy by 7--18pp, showing that cultural-geographic knowledge exists but is not reliably activated through visual input.}
\label{tab:text_only_probe}
\end{table}

\section{Case Study Details}
\label{sec:case_study_appendix}

The two tasks in Fig.~\ref{fig:case} illustrate the multi-step reasoning required by \BenchName:

\paragraph{Task 1: Semantic Disambiguation.} The model must disambiguate ``English Cheese Tea Treats'', where ``cheese tea'' evokes East Asian beverage culture while ``treats'' suggests solid desserts. The model decomposes the query (``English'' $\to$ origin; ``Cheese Tea'' $\to$ beverage; ``Treats'' $\to$ dessert confusion), notes Chinese watermarks contradicting the Anglophone framing, and aligns against visual evidence (berries, cream, ramekins $\to$ crumble/dessert) to conclude ``No.''

\paragraph{Task 2: Visual Abstraction.} Both images depict Hasselback potatoes with characteristic fan-cut geometry, but differ in plating and garnish. The model must abstract away surface differences and focus on core identity markers: cut pattern, primary ingredient, and cooking method. It then recognizes both as Western/European-inspired preparations, demonstrating categorical equivalence despite stylistic variation.

\end{document}